\documentclass[runningheads]{llncs}

\usepackage[final,year=2026,ID=11095]{eccv}

\usepackage{eccvabbrv}

\usepackage{graphicx}
\usepackage{booktabs}

\usepackage[accsupp]{axessibility}  
\usepackage{multirow}

\usepackage{hyperref}

\usepackage{orcidlink}

\begin{document}

\title{Second-Order Multi-Level Variance Correction for Modality Competition in Multimodal Models} 

\titlerunning{Abbreviated paper title}


\author{Yishun Lu and Wes Armour}


\institute{University of Oxford, Oxford, United Kingdom \\
\email{yishun.lu@eng.ox.ac.uk, wes.armour@oerc.ox.ac.uk}\\}

\maketitle

\begin{abstract}
Autoregressive next-token training offers a unified formulation for image generation and text understanding, but it also creates strong modality competition that destabilizes optimization and limits large-batch scaling. We show that first-order optimizers such as AdamW are vulnerable to cross-modality gradient heterogeneity, while second-order preconditioning, particularly SOAP, provides a more stable basis for multimodal alignment. Building on this insight, we propose \emph{ML-FOP-SOAP}, a second-order optimization framework with Multi-Level Variance Correction. Our Fisher-Orthogonal Projection suppresses variance-induced modality conflicts, reducing the trade-off between visual generation and textual understanding. To make this practical under large gradient accumulation, we introduce a hierarchical folding strategy that captures fine-grained variance with low micro-step overhead. Experiments on Janus and Emu3 show consistent gains across both modalities and stable training at batch size 8192. Compared with AdamW, our method improves sample efficiency by up to $1.4\times$ and accelerates wall-clock training by up to $1.5\times$, offering a robust optimizer for scaling multimodal foundation models.

\end{abstract}

\section{Introduction}

In recent years, unifying image generation and text understanding under the autoregressive next-token prediction paradigm has become a major trend in the development of vision-language foundation models. However, this deep multi-modal integration introduces a severe modality competition problem during pretraining, which often makes joint training highly unstable and convergence extremely slow. To address this challenge, most existing approaches in the literature primarily focus on model architecture. For example, researchers have attempted to physically disentangle modality-specific representations by designing decoupled visual encoders, introducing multi-way attention mechanisms, or constructing dedicated alignment modules~\cite{team2024chameleon,wang2026multimodalemu3,wu2025janus,liu2023visual,bai2023qwenvlversatilevisionlanguagemodel}. 

Beyond architectural modifications, existing efforts to alleviate modality conflict, such as dynamic loss weighting, curriculum scheduling, and gradient balancing~\cite{chen2018gradnorm,dai2023instructblip,kontras2024improving,qian2025dyncim}, remain largely external to the optimizer. In practice, parameter updates are still governed by first-order methods like AdamW~\cite{loshchilov2017decoupled}, which rely on diagonal moment estimates. These methods fail to model the underlying local curvature of the loss surface or the geometric coupling between modalities.

This reliance on first-order optimization becomes a critical bottleneck because multi-modal training induces gradient estimators with vastly different noise patterns and curvature sensitivities~\cite{liang2022mind}. Specifically, visual objectives often involve high-dimensional and spatially diffuse prediction errors, whereas textual objectives are typically concentrated on a more compact semantic manifold ~\cite{yu2020gradient}. This geometric mismatch creates a complex cross-modal covariance structure. Consequently, under standard diagonal preconditioning, modalities associated with larger gradient variance (vision) exert a disproportionate influence on the update trajectory, suppressing lower-variance but semantically dense signals (text)~\cite{peng2022balanced,wilson2017marginal}. 

To resolve this variance conflict at the optimization level, the recently proposed Fisher-Orthogonal Projection (FOP)~\cite{lu2025beyond} accelerates convergence by projecting gradients in Fisher geometry. However, its original reliance on K-FAC~\cite{martens2015optimizing} is memory-prohibitive for the massive vocabularies and wide hidden layers of modern foundation models. To address this, we leverage structured two-dimensional (2D) tensor preconditioners. Interestingly, our empirical analysis reveals that while standard methods like Shampoo~\cite{shi2023distributed} are easily destabilized by early-stage visual outliers, the SOAP optimizer~\cite{vyas2024soap,abreu2025potential} provides a highly robust, memory-efficient surrogate for local optimization metrics. Yet, even with SOAP, we observe a lingering trade-off where textual understanding is inadvertently sacrificed to accommodate high-entropy visual noise. Thus, integrating FOP with SOAP becomes essential to fully neutralize these cross-modality conflicts.

Furthermore, scaling these unified models to ultra-large global batch sizes (e.g., 8192) requires massive gradient accumulation across numerous micro-steps. Applying variance correction only at the coarse boundaries of a long accumulation window fails to preserve the fine-grained gradient discrepancies arising within the window. Inspired by the telescoping coarse-to-fine viewpoint of the Multi-Level Monte Carlo (MLMC) paradigm~\cite{giles2015multilevel}, we propose a novel Multi-level Hierarchical Gradient Folding mechanism via Fisher-Orthogonal Projection (ML-FOP) . Instead of classical scalar expectation estimation, our approach recursively fuses computationally lightweight cumulative snapshots to preserve intra-step multi-scale variance to achieve multi-level variance correction.

Through this methodology, we present a new optimization framework that natively mitigates modality competition and enables the training of unified autoregressive models with ultra-large global data batches, dramatically improving hardware utilization on modern GPU clusters.

Specifically, our contributions are as follows:

\begin{itemize}
    \item \textbf{Second-Order Optimization for Modality Competition.} We systematically investigate the optimization dynamics of unified multi-modal autoregressive models. We reveal that while first-order methods (AdamW~\cite{loshchilov2017decoupled}) and standard uncentered second-order methods (Shampoo~\cite{shi2023distributed}) struggle with extreme visual gradient outliers, the SOAP optimizer~\cite{vyas2024soap} inherently mitigates this cross-modality heterogeneity, establishing a robust foundation for multi-modal alignment. Compared to AdamW, our ML-FOP-SOAP framework delivers a $1.4\times$ gain in sample efficiency and a $1.5\times$ acceleration in end-to-end wall-clock time, uniquely preserving robust performance across both modalities.

    \item \textbf{Variance Control via Fisher-Orthogonal Projection (FOP).} Building upon the SOAP baseline, we introduce a geometry-aware FOP mechanism. We demonstrate that projecting gradient differences orthogonally to the primary descent trajectory actively neutralizes modality competition. This effectively eliminates the conventional trade-off between textual understanding and visual generation, yielding strict Pareto improvements across both tasks.

    \item \textbf{Efficient Multi-Level FOP for Massive Accumulation.} To make FOP tractable under massive gradient accumulation, we propose an efficient Multi-Level hierarchical folding mechanism (ML-FOP). This dyadic telescoping strategy captures fine-grained, cross-modality variance across accumulation segments without the prohibitive computational wall-clock overhead of applying projections at every micro-step.

    \item \textbf{Ultra-Large-Batch multi-modal Alignment Benchmark.} We validate our framework on representative unified architectures (Janus and Emu3) scaled on massive datasets (LLaVA-3M and 12M~\cite{lmmslab_llava_recap_cc3m_2024,lmmslab_llava_recap_cc12m_2024}). We successfully scale the global batch size up to 8192, a regime where existing optimizers suffer catastrophic degradation. 
\end{itemize}
\section{Related Work}

\subsection{Native Unified Vision-Language Models} 
The dominant paradigm in vision-language foundation models has recently shifted towards unifying image and text modalities within a single autoregressive next-token prediction framework~\cite{team2024chameleon,wu2025janus,wang2026multimodalemu3}. Unlike earlier compositional architectures that connect a pre-trained vision encoder (e.g., CLIP) to a Large Language Model via projection layers~\cite{liu2023visual,bai2023qwenvlversatilevisionlanguagemodel}, native autoregressive models map all continuous visual signals into discrete tokens, treating both modalities equally. While this architectural convergence enables seamless generation and understanding, it inevitably exacerbates modality competition during joint pre-training. Our work does not alter these advanced model architectures; rather, we provide the optimization infrastructure necessary to stabilize their training and scale them to unprecedented batch sizes.

\subsection{Modality Competition and Multitask Optimization} 
The phenomenon of modality conflict is closely related to gradient interference in multitask learning. Traditional approaches to alleviate this include dynamic loss weighting~\cite{chen2018gradnorm}, gradient projection techniques, and curriculum-based data scheduling~\cite{dai2023instructblip,kontras2024improving,qian2025dyncim}. However, these methods operate largely external to the optimizer itself. Parameter updates in foundation models are still universally governed by first-order optimizers like AdamW~\cite{loshchilov2017decoupled}. AdamW relies on diagonal preconditioning, which independently scales coordinates but fundamentally fails to capture the complex, non-diagonal cross-modal covariance structure and the distinct local geometry induced by high-variance visual objectives and dense textual signals~\cite{liang2022mind}.

\subsection{Efficient Second-Order Preconditioning}
To address the limitations of first-order methods, second-order optimization explicitly models the local curvature of the loss surface. The gold standard for this is Natural Gradient Descent (NGD), which preconditions the Euclidean gradient $g$ using the inverse of the Fisher Information Matrix (FIM), $F$, yielding the optimal update direction $\tilde{g} = F^{-1}g$. However, computing and inverting the exact FIM is fundamentally intractable for deep neural networks. Previous methods often employ explicit approximations, such as K-FAC~\cite{martens2015optimizing}, to make the inversion feasible. The recently proposed Fisher-Orthogonal Projection (FOP)~\cite{lu2025beyond} successfully leverages K-FAC to accelerate multi-modal convergence. Unfortunately, explicitly modeling curvature via K-FAC still incurs catastrophic memory overhead when applied to modern VLMs with massive vocabularies and wide hidden layers.

To scale geometry-aware optimization, we turn to lightweight tensor preconditioners like Shampoo~\cite{shi2023distributed} and SOAP~\cite{vyas2024soap,abreu2025potential}, which construct preconditioners based on empirical second-order gradient statistics. Crucially, while standard Shampoo utilizes uncentered second moments that are highly vulnerable to early-stage visual gradient outliers, SOAP projects gradients into a skewed eigenbasis and applies coordinate-wise adaptive normalization. This provides a highly robust, memory-efficient surrogate for the natural gradient: $\mathcal{P}(g) \approx F^{-1}g$. By replacing explicit Fisher inversions with SOAP, our framework amortizes the prohibitive memory overhead while capturing the local geometry necessary for multi-modal training.

\section{Methodology}
\label{sec:methodology}

\subsection{Mathematical Formulation of Modality Competition}
\label{sec:modality_competition}

Our goal is to formalize modality competition in unified autoregressive multi-modal models, such as Janus ~\cite{wu2025janus} and Emu3 \cite{wang2026multimodalemu3}, as an optimization imbalance induced by heterogeneous stochastic gradients. Consider a model parameterized by $\theta$, trained on both image and text data. The joint objective can be written as
\begin{equation}
\mathcal{L}(\theta)
=
\mathbb{E}_{x \sim \mathcal{D}_{\mathrm{img}}}\left[L_{\mathrm{img}}(\theta; x)\right]
+
\mathbb{E}_{y \sim \mathcal{D}_{\mathrm{text}}}\left[L_{\mathrm{text}}(\theta; y)\right]
\label{eq:joint_multimodal_objective}
\end{equation}

In stochastic optimization, the full gradient $\nabla \mathcal{L}(\theta)$ is approximated using minibatches $\mathcal{B}_{\mathrm{img}} \subset \mathcal{D}_{\mathrm{img}}$ and $\mathcal{B}_{\mathrm{text}} \subset \mathcal{D}_{\mathrm{text}}$. The practical update is therefore based on the stochastic gradient:

\begin{equation}
g
=
g_{\mathrm{img}} + g_{\mathrm{text}}
=
\frac{1}{|\mathcal{B}_{\mathrm{img}}|}
\sum_{x \in \mathcal{B}_{\mathrm{img}}}
\nabla_{\theta} L_{\mathrm{img}}(\theta; x)
+
\frac{1}{|\mathcal{B}_{\mathrm{text}}|}
\sum_{y \in \mathcal{B}_{\mathrm{text}}}
\nabla_{\theta} L_{\mathrm{text}}(\theta; y)
\label{eq:stochastic_joint_gradient}
\end{equation}
which serves as a Monte Carlo estimator of the full gradient $\nabla \mathcal{L}(\theta)$. The key difficulty is that the image and text components generally induce gradient estimators with substantially different noise scales and covariance structures. Let
$
\Sigma_{\mathrm{img}}
=
\mathrm{Cov}(g_{\mathrm{img}})
$
,
$
\Sigma_{\mathrm{text}}
=
\mathrm{Cov}(g_{\mathrm{text}})
$.
In many multi-modal training scenarios, it can be observed that a strong imbalance between these covariance scales, often reflected by
$
\mathrm{Tr}\!\left(\Sigma_{\mathrm{img}}\right)
\gg
\mathrm{Tr}\!\left(\Sigma_{\mathrm{text}}\right)
$.
Examples can be seen in the supplementary material.
This imbalance does not merely affect the magnitude of the stochastic gradient, but also its effective optimization geometry in the shared parameter space. Under first-order adaptive optimizers such as AdamW ~\cite{loshchilov2017decoupled}, the update is constructed from per-parameter first and second moment estimates. Abstracting away bias correction for clarity, the update takes the form
$d_t=\frac{m_t}{\sqrt{v_t} + \epsilon}$,
where $m_t$ and $v_t$ denote the exponential moving averages of the gradient and squared gradient. Since this update rule relies on diagonal preconditioning, it can only rescale coordinates independently and does not explicitly model cross-parameter coupling or curvature interactions across modalities ~\cite{abreu2025potential}. As a result, when one modality contributes disproportionately high gradient variance, its noisy directions may dominate the optimization trajectory, while weaker but semantically informative signals from another modality are underutilized. 

\subsection{Fisher Information for Curvature Awareness}
\label{sec:fisher_necessity}

To resolve modality conflict induced by Euclidean first-order updates, we seek an optimization geometry that reflects the intrinsic structure of the model distribution rather than raw coordinate-wise gradient magnitudes. Natural gradient descent ~\cite{pascanu2013revisiting} provides such a perspective by measuring parameter perturbations through the change they induce in the predictive distribution. Specifically, for a small parameter step $\Delta\theta$, the Kullback--Leibler divergence between the model before and after the update admits the local approximation
\begin{equation}
D_{\mathrm{KL}}\!\left(p_{\theta} \,\|\, p_{\theta+\Delta\theta}\right)
\approx
\frac{1}{2}\Delta\theta^{\top} F \Delta\theta
\label{eq:kl_local_fisher_method}
\end{equation}
where, $F$ is the Fisher information matrix, defined as
\begin{equation}
F
=
\mathbb{E}_{z \sim p(\cdot \mid \theta)}
\left[
\nabla \log p(z \mid \theta)\nabla \log p(z \mid \theta)^{\top}
\right]
\label{eq:fisher_definition_method}
\end{equation}

Equation~\eqref{eq:kl_local_fisher_method} shows that the Fisher information matrix defines the local metric of the statistical manifold. Unlike diagonal first-order preconditioners in AdamW, $F$ captures structured curvature information and parameter coupling, thereby providing a geometry-aware notion of distance in parameter space.

This is especially important in multi-modal training. High-variance visual gradients may correspond to directions that are statistically noisy or locally flat, while lower-variance textual gradients may align with sharper and more informative directions for joint generalization. Preconditioning the gradient by $F^{-1}$ yields the natural gradient update
\begin{equation}
d_{\mathrm{NGD}}
=
F^{-1} g
\label{eq:natural_gradient_update_method}
\end{equation}

Under this transformation, updates are no longer determined solely by Euclidean scale, but are normalized with respect to the local geometry of the model distribution. Consequently, directions associated with large stochastic variance can be damped when they are not supported by informative curvature, whereas lower-variance but high-curvature directions can be preserved or amplified. This provides the mathematical motivation for introducing Fisher-aware optimization in order to balance multi-modal training dynamics. The details of proof are shown in Supplementary section \ref{app:theory_fisher_modality_balancing}.

\subsection{Tractable Curvature via Tensor Preconditioners}
\label{sec:tensor_preconditioners_limitations}

To make curvature-aware optimization tractable for large multi-modal autoregressive models, we must address the prohibitive computational and memory costs of the exact Fisher Information Matrix $F$. For a neural network with $N$ parameters, the Fisher matrix $F \in \mathbb{R}^{N \times N}$ requires $\mathcal{O}(N^3)$ for explicit inversion, which is fundamentally infeasible for models with billions of parameters. 

Structured tensor preconditioners, such as Shampoo~\cite{shi2023distributed} and SOAP~\cite{vyas2024soap}, provide a scalable approximation to the geometry induced by $F$. Consider a parameter tensor whose gradient is represented as a matrix $G \in \mathbb{R}^{m \times n}$, and let $g = \mathrm{vec}(G)$ be its vectorized form. These methods approximate the local curvature by assuming a Kronecker-factored structure of the second-moment matrix:
\begin{equation}
\hat{F} \approx L \otimes R
\label{eq:fisher_kronecker_approx}
\end{equation}
where the left and right factors $L \in \mathbb{R}^{m \times m}$ and $R \in \mathbb{R}^{n \times n}$ capture the correlations along the respective tensor dimensions:
\begin{equation}
L \approx \mathbb{E}[G G^{\top}], \qquad R \approx \mathbb{E}[G^{\top} G]
\label{eq:shampoo_factors}
\end{equation}

Crucially, unlike Natural Gradient Descent (NGD) which typically applies the inverse Fisher $F^{-1}$, tensor preconditioners like SOAP aim to approximate the \textit{inverse square-root} of the Fisher matrix, $F^{-1/2}$, to achieve adaptive coordinate-wise scaling in the spectral domain. Under the property $(L \otimes R)^{-1/2} = L^{-1/2} \otimes R^{-1/2}$, the preconditioned update can be computed without materializing the full $N \times N$ matrix. In practice, to maintain symmetric preconditioning for the matrix $G$, the update is applied as:
\begin{equation}
\mathcal{P}(g) = \mathrm{vec}\left( L^{-1/4} G R^{-1/4} \right)
\label{eq:inverse_fisher_application}
\end{equation}
where $L^{-1/4}$ and $R^{-1/4}$ are computed via eigendecomposition (as in SOAP) or iterative methods (as in Shampoo). Equation~\eqref{eq:inverse_fisher_application} is mathematically equivalent to applying $(R^{1/4} \otimes L^{1/4})^{-1} g$, providing a memory-efficient surrogate for the geometry-aware update.

Despite this efficiency, applying such preconditioning directly to the averaged batch gradient $\bar{g}$ still suffers from severe modality competition. The fundamental limitation lies in the sequential nature of the standard update:
\begin{equation}
\Delta \theta_{\mathrm{SOAP}} \propto \mathcal{P}(\bar{g}), \quad \text{where} \quad \bar{g} = \frac{1}{K}\sum_{i=1}^{K} g_i
\label{eq:soap_update_avg}
\end{equation}
During the averaging step, fine-grained gradient discrepancies between micro-batches are smoothed out. In multi-modal training, where image and text samples produce gradients with vastly different noise scales and directions, the averaged vector $\bar{g}$ becomes biased towards the modality with higher variance (typically vision). Consequently, the preconditioner $\mathcal{P}(\cdot)$ only adjusts the coarse, batch-level geometry of an already biased signal; it cannot recover the modality-specific informative directions that were neutralized during the initial averaging process. This motivates the need for variance-aware correction before the preconditioning stage.

\subsection{Variance Control via Fisher-Orthogonal Projection}
\label{sec:fop_variance_control}

To recover the critical intra-batch variance information that is prematurely lost when simply averaging gradients, we introduce Fisher-Orthogonal Projection (FOP) on gradient differences. During gradient accumulation, let $g_1$ and $g_2$ denote two stochastic gradients computed from distinct micro-batches. We define their average and difference as:
\begin{equation}
g_{\mathrm{avg}} = \frac{1}{2}(g_1 + g_2), \qquad g_{\mathrm{diff}} = g_1 - g_2
\label{eq:avg_and_diff}
\end{equation}
Here, $g_{\mathrm{diff}}$ inherently captures the micro-level variance and heterogeneity within the accumulation window. In the context of multi-modal alignment, it explicitly encodes the optimization conflict between processing an image-centric micro-batch and a text-centric micro-batch.

Instead of explicitly forming the prohibitive Fisher matrix $F$ to perform geometric projections, we construct a lightweight forward metric proxy $\mathcal{M}(v) \approx Fv$ directly from the internal states of the tensor preconditioners (e.g., Shampoo or SOAP). Using this computationally efficient proxy, we extract a Fisher-orthogonal residual $g_{\mathrm{diff}}^{\perp}$ by projecting $g_{\mathrm{diff}}$ away from the mean descent direction $g_{\mathrm{avg}}$ in the Riemannian manifold:
\begin{equation}
g_{\mathrm{diff}}^{\perp} = g_{\mathrm{diff}} - \left( \frac{\langle g_{\mathrm{diff}}, \mathcal{M}(g_{\mathrm{avg}}) \rangle}{\langle g_{\mathrm{avg}}, \mathcal{M}(g_{\mathrm{avg}}) \rangle + \epsilon} \right) g_{\mathrm{avg}}
\label{eq:fisher_orthogonal_residual_main}
\end{equation}
where $\epsilon$ is a small numerical stabilizer. By construction, this residual satisfies $\langle g_{\mathrm{diff}}^{\perp}, \mathcal{M}(g_{\mathrm{avg}}) \rangle \approx \langle g_{\mathrm{diff}}^{\perp}, F g_{\mathrm{avg}} \rangle = 0$. It perfectly isolates the pure cross-modality variance without interfering with the primary optimization trajectory.

The final combined gradient direction is then constructed by augmenting the average gradient with this variance-aware correction:
\begin{equation}
g_{\mathrm{comb}} = g_{\mathrm{avg}} + \beta g_{\mathrm{diff}}^{\perp}
\label{eq:combined_direction}
\end{equation}
where $\beta$ is an adaptive mixing coefficient. This combined direction is subsequently mapped back to the parameter space using the preconditioner $\mathcal{P}(\cdot)$ to obtain the final update step:
\begin{equation}
d_{\mathrm{FOP}} = F^{-1} g_{\mathrm{comb}}
\label{eq:fop_update}
\end{equation}

This formulation elegantly injects variance control into the preconditioned subspace without breaking the tractable $\mathcal{O}(m^3 + n^3)$ complexity bound. The detailed step-by-step mathematical derivation, including the computation of the optimal adaptive coefficient $\beta$, is provided in Supplementary material Section ~\ref{app:fop_derivation}.

Standard SOAP or Shampoo updates are inherently restricted to the one-dimensional subspace defined by the batch mean, yielding an update $d_{\mathrm{SOAP}} \approx F^{-1} g_{\mathrm{avg}}$. By contrast, FOP expands the search space by injecting the Fisher-orthogonal variance residual $g_{\mathrm{diff}}^{\perp}$. As we show in Supplementary material Section ~\ref{app:quadratic_surrogate}, under a local quadratic surrogate objective $J(d)$ governing the natural gradient descent, the FOP update satisfies an inequality:
\begin{equation}
J_{\mathrm{FOP}}(\beta) < J_{\mathrm{SOAP}}, \quad \forall \beta \neq 0
\label{eq:fop_superiority}
\end{equation}

\subsection{Multi-Scale Variance Correction via MLMC-Inspired Hierarchical Folding}
\label{sec:mlmc_inspired_folding}

To scale multi-modal autoregressive models to ultra-large batch sizes (e.g., 8192), training must rely on extensive gradient accumulation across $K$ micro-steps. Applying standard FOP only once, using coarse summaries from the start and end of a long accumulation window, would fail to explicitly preserve intermediate fine-scale gradient discrepancies arising within the window.

To address this, we borrow the telescoping coarse-to-fine viewpoint of the Multi-Level Monte Carlo (MLMC) paradigm~\cite{giles2015multilevel}. In classical computational statistics, MLMC achieves optimal variance reduction by telescoping an expectation across a hierarchy of discretization resolutions, effectively combining coarse global estimates with fine local corrections. 

Unlike classical MLMC, our objective here is not to estimate a scalar expectation via unbiased multi-level sampling, but rather to preserve the rich, within-step variance structure of gradients during massive accumulation. We propose a \textbf{Multi-level Hierarchical Gradient Folding} mechanism. We discretize the accumulation window into a dyadic hierarchy of $L$ levels, where the snapshot intervals $L_j \in \{1, 2, 4, 8, \dots, K\}$. 

Instead of discarding intermediate gradients, we record computationally lightweight cumulative average snapshots:
\begin{equation}
\bar{g}_{1:k} = \frac{1}{k}\sum_{i=1}^k g_i
\label{eq:cumulative_snapshots}
\end{equation}
By simple algebra on cumulative averages, we reconstruct the missing segment averages (the segment-level averages), bridging adjacent dyadic levels without explicitly storing every micro-batch:
\begin{equation}
\bar{g}_{L_{j-1}+1 : L_j} = \frac{L_j \bar{g}_{1:L_j} - L_{j-1} \bar{g}_{1:L_{j-1}}}{L_j - L_{j-1}}
\label{eq:segment_average}
\end{equation}

By treating these recovered segment averages as multi-scale summaries whose discrepancies encode variance across the accumulation hierarchy, we perform a recursive, bottom-up folding process using our FOP operator:
\begin{equation}
z_0 = \bar{g}_{1:1}, \qquad z_j = \mathrm{FOP}_{\mathrm{accum}}(z_{j-1}, \bar{g}_{L_{j-1}+1:L_j})
\label{eq:mlmc_fop_recursion}
\end{equation}
where $\mathrm{FOP}_{\mathrm{accum}}(\cdot, \cdot)$ combines the previous folded state with the recovered segment average through Fisher-aware residual filtering.

\section{Experimental Setup}

\label{sec:model_and_dataset}
\subsection{Model and dataset}
To systematically evaluate the efficacy of our hierarchical folding optimization, we conduct experiments on two representative unified multi-modal autoregressive architectures: Janus~\cite{wu2025janus} and Emu3~\cite{wang2026multimodalemu3}. While both models utilize an autoregressive framework for unified multi-modal alignment, they adopt fundamentally distinct visual encoding paradigms. Emu3 relies on a purely discrete tokenizer, fully converting continuous visual signals into a discrete space to share Fa single, unified vocabulary with text. In contrast, Janus employs a decoupled visual encoding strategy: it extracts continuous semantic features (via SigLIP ~\cite{zhai2023sigmoidlosslanguageimage}) for multi-modal understanding, while utilizing a VQ tokenizer ~\cite{oord2018neuraldiscreterepresentationlearning} to map images into discrete IDs specifically for visual generation. To accommodate their respective architectural designs and spatial resolution requirements, we configure the maximum training sequence length to 1024 for the Janus variants and 5120 for the Emu3 variant. Furthermore, rather than directly adopting the massive parameter scales from their original papers (which typically range from 7B to over 8B parameters ), we construct three scaled-down, intermediate-capacity variants: \textbf{Janus-400M}, and \textbf{Emu3-600M}.

The primary motivation for this downscaling is computational tractability and ablation rigor. Investigating the intricate optimization dynamics of modality competition under ultra-large batch sizes (e.g., 8192) requires extensive hyperparameter tuning and hundreds of gradient accumulation steps. Evaluating full-scale 8B models under these extreme accumulation settings is computationally prohibitive. By systematically scaling down the network capacity while maintaining the core architectural proportions, these intermediate variants perfectly preserve the severe cross-modality gradient heterogeneity observed in full-scale models. This allows us to rigorously and efficiently ablate our multi-scale variance control mechanisms. The exact scaling configurations, layer depths, and parameter counts for each variant are detailed in Supplementary material section ~\ref{app:model_configurations}.

To thoroughly evaluate the robustness and efficacy of our optimization methods under realistic, large-scale conditions, we construct a massive multi-modal pre-training corpus by combining the widely adopted \textbf{LLaVA-3M} and \textbf{LLaVA-12M}  datasets ~\cite{lmmslab_llava_recap_cc12m_2024,lmmslab_llava_recap_cc3m_2024,liu2023llavapluslearningusetools}. By jointly training our scaled-down autoregressive models on this combined dataset, we expose the optimization trajectory to highly diverse and extensive image-text pairs. This comprehensive data scale is specifically designed to induce severe cross-modality gradient heterogeneity, providing a rigorous testbed for validating our multi-scale variance control under massive gradient accumulation settings.

\subsection{Implementation Details}

To ensure balanced multi-modal alignment, both architectures are jointly trained on Image-to-Text (I2T) and Text-to-Image (T2I) tasks with a 50\%/50\% probability split. However, the specific mixing strategies are tailored to their respective architectural paradigms. For \textbf{Janus}, we utilize step-level synchronous task routing across distributed ranks. In each training step, a specific task (understanding or generation) is randomly sampled and synchronized across all ranks to maintain consistent gradient directions for the decoupled encoders. In contrast, \textbf{Emu3} leverages its unified next-token prediction framework to implement sample-level mixing within each batch. To further enhance bidirectional alignment and prevent positional bias, we randomly permute the relative order of image tokens and text captions within each micro-batch.

All experiments are executed on a high-performance cluster equipped with \textbf{AMD MI300X} GPUs, each providing 192GB of VRAM. To handle the memory demands of large-scale pre-training, especially for Emu3's extended sequence length of 5120 tokens, we employ gradient checkpointing across all model variants. The micro-batch size per GPU is configured as 64 for \textbf{Janus-400M} and 8 for \textbf{Emu3-600M}. We utilize gradient accumulation to reach the target global batch size, and to maintain stable optimization dynamics, we strictly adhere to the square-root learning rate scaling rule ($\eta_{\mathrm{scaled}} = \eta_{\mathrm{base}} \times \sqrt{B_{\mathrm{total}} / 1024}$). 

For hyperparameter tuning, we conduct a systematic grid search to identify the optimal peak learning rate for each optimizer and model variant. Regarding the learning rate schedules, we use a 10\% warmup ratio by default. We uses cosine decay with a minimum LR floor of 1e-6. Crucially, Janus defaults to freezing all vision-related modules, tightly focusing the optimization on the autoregressive backbone and naturally accounting for the gap between its trainable and total parameter counts.

\section{Experiments}

\begin{figure}
    \centering
    \includegraphics[width=1\linewidth]{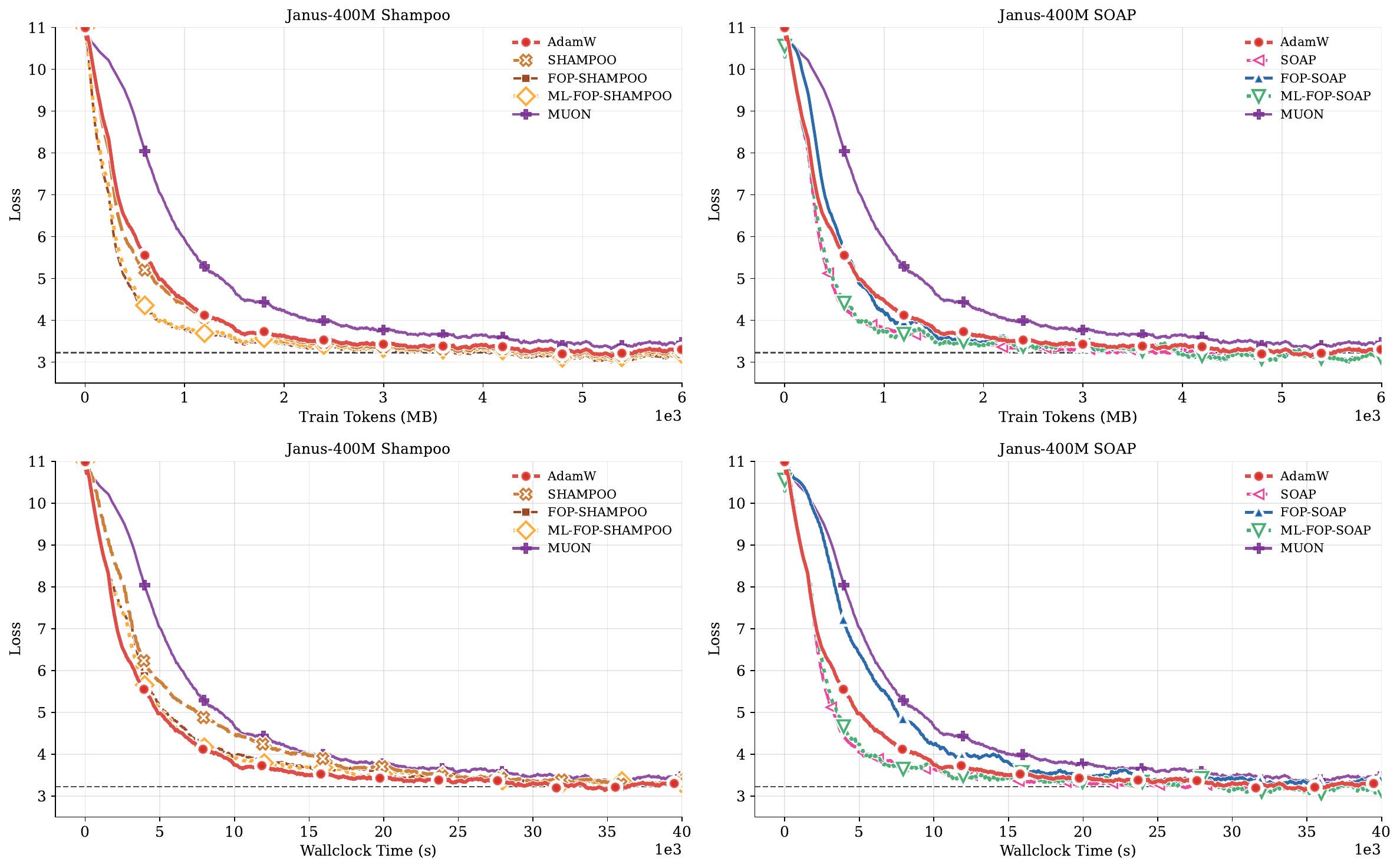}
    \caption{2x2 train-loss comparison for pretraining Janus-400M. Left column: SHAMPOO family; right column: SOAP family. Top row: loss vs trained tokens; bottom row: loss vs wallclock time.}
    \label{fig:loss landscape}
\end{figure}

The overall training efficiency and convergence behaviors of the proposed optimization strategies are first evaluated under a baseline batch size of 1024. In Figure~\ref{fig:loss landscape}, a comparative analysis of the training loss for the Janus-400M architecture is presented across the SHAMPOO (left column) and SOAP (right column) optimizer families. To rigorously define the optimization plateau, the dashed horizontal lines indicate the convergence thresholds established by extending the AdamW baseline training for $20\times$ the standard token volume according to a ``Chinchilla-optimal'' scaling law~\cite{hoffmann2022trainingcomputeoptimallargelanguage}. 

Under this regime, the SHAMPOO family underperforms the AdamW baseline. Early-stage vision-text gradient outliers heavily skew SHAMPOO's uncentered second-moment matrices. Conversely, SOAP mitigates this by projecting gradients into the skewed eigenbasis and applying coordinate-wise adaptive normalization, effectively insulating the network from divergence. Consequently, standard SOAP reaches the convergence threshold at merely 3.2B tokens—a $\sim 1.4\times$ sample efficiency improvement over AdamW (4.5B tokens), while MUON entirely fails to approach it.


\begin{table}[htbp]
\centering
\caption{Cross-architecture throughput and memory comparison between Janus-400M and Emu3-600M under global batch sizes (BS) of 1024 and 8192. For each architecture, the results are split into two sub-rows: the first row reports the average time per global training step (seconds), and the second row reports the peak VRAM consumption (GB).}
\label{tab:final_efficiency_comparison}
\resizebox{\linewidth}{!}{
\begin{tabular}{llcccccccc|cccccccc}
\toprule
Batch Size & & \multicolumn{8}{c|}{\textbf{BS = 1024}} & \multicolumn{8}{c}{\textbf{BS = 8192}} \\
\cmidrule(lr){3-10} \cmidrule(lr){11-18}
\textbf{Arch} & \textbf{Metric} & AdamW & MUON & Sham. & F-Sh. & ML-F-Sh. & SOAP & F-So. & ML-F-So. & AdamW & MUON & Sham. & F-Sh. & ML-F-Sh. & SOAP & F-So. & ML-F-So. \\
\midrule
\multirow{2}{*}{\textbf{Janus}} & Time & 7.76 & 7.85 & 12.93 & 15.64 & 13.87 & 8.40 & 12.22 & 8.62 & 60.94 & 60.90 & 59.91 & 74.48 & 114.32 & 65.53 & 78.93 & 67.43 \\
 & VRAM & 121.86 & 121.48 & 128.49 & 126.92 & 126.93 & 129.39 & 127.28 & 131.65 & 121.86 & 121.48 & 128.49 & 126.92 & 126.93 & 129.39 & 127.28 & 131.65 \\
\midrule
\multirow{2}{*}{\textbf{Emu3}} & Time & 236.91 & 252.51 & N/A & N/A & N/A & 250.63 & 273.56 & 259.67 & 1859.13 & 1949.13 & N/A & N/A & N/A & 1945.83 & 2135.59 & 1971.47 \\
 & VRAM & 144.38 & 139.91 & N/A & N/A & N/A & 145.61 & 146.77 & 151.08 & 144.38 & 139.91 & N/A & N/A & N/A & 145.61 & 146.77 & 151.08 \\
\bottomrule
\end{tabular}
}
\end{table}

Crucially, applying the Fisher-Orthogonal Projection (FOP) to the SOAP optimizer amplifies stability. By neutralizing variance-induced cross-modality conflicts, FOP-SOAP and ML-FOP-SOAP break through vanilla SOAP's plateau to achieve lower global losses. Furthermore, the Multi-Level (ML) mechanism successfully amortizes FOP's projection latency. As a result, ML-FOP-SOAP hits the target threshold in around 3.1B tokens (a $\sim 1.4\times$ efficiency gain over AdamW) and requires around 5.5 hours of wall-clock time. This yields an end-to-end actual training acceleration of $\sim 1.5\times$ compared to AdamW's 8.3 hours. 

These findings substantiate that robust second-order preconditioners establish a critical multi-modal foundation, while FOP-based variance control unlocks deeper convergence. Given SHAMPOO's vulnerability to early-stage outliers, subsequent analyses will focus exclusively on the SOAP family.

\begin{figure}
    \centering
    \includegraphics[width=1\linewidth]{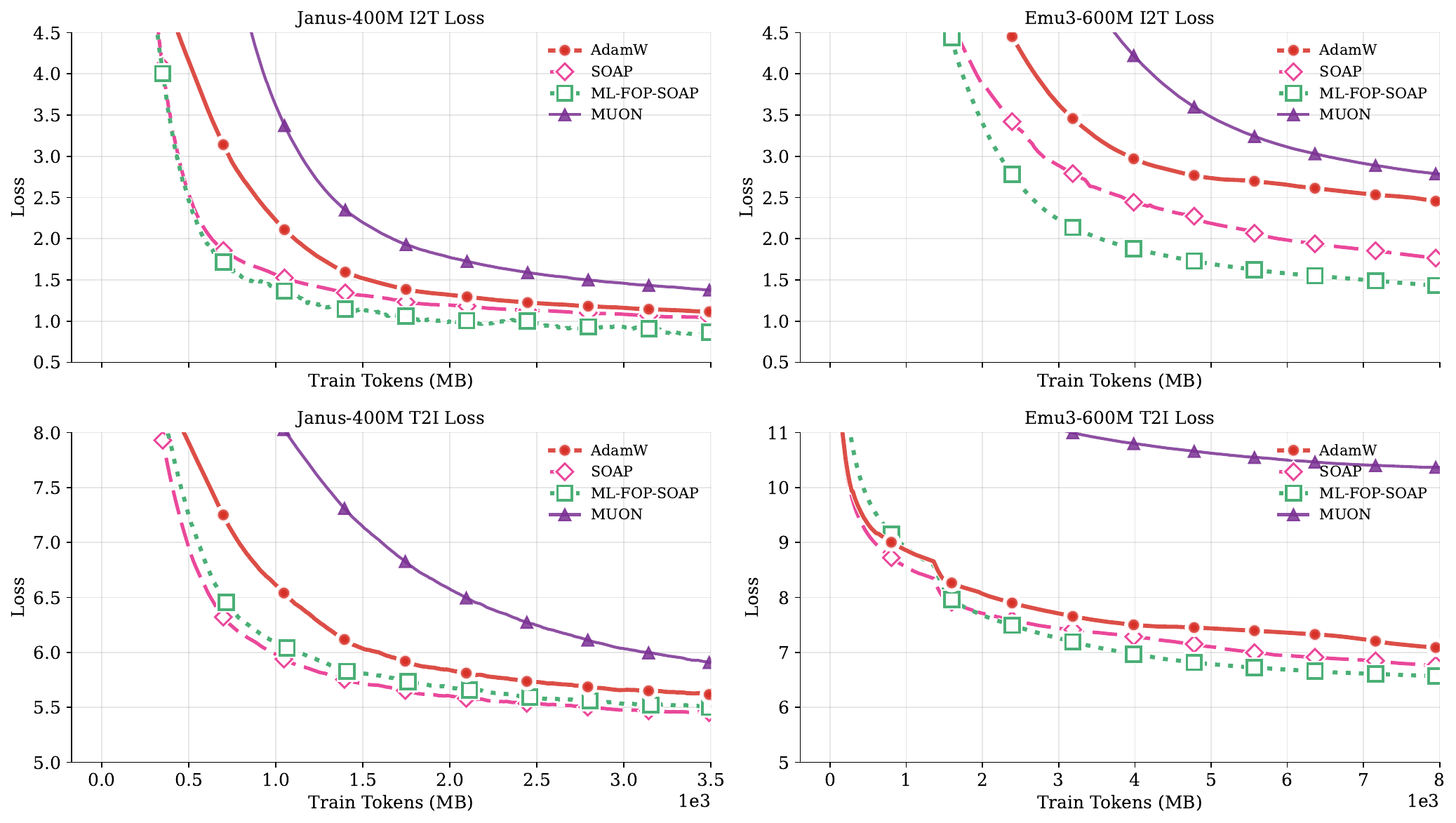}
    \caption{Decoupled training loss curves for image-to-text (I2T) and text-to-image (T2I) tasks across Janus-400M (left) and Emu3-600M (right) architectures.}
    \label{fig:Decoupled training loss curves}
\end{figure}

\begin{figure}
    \centering
    \includegraphics[width=1\linewidth]{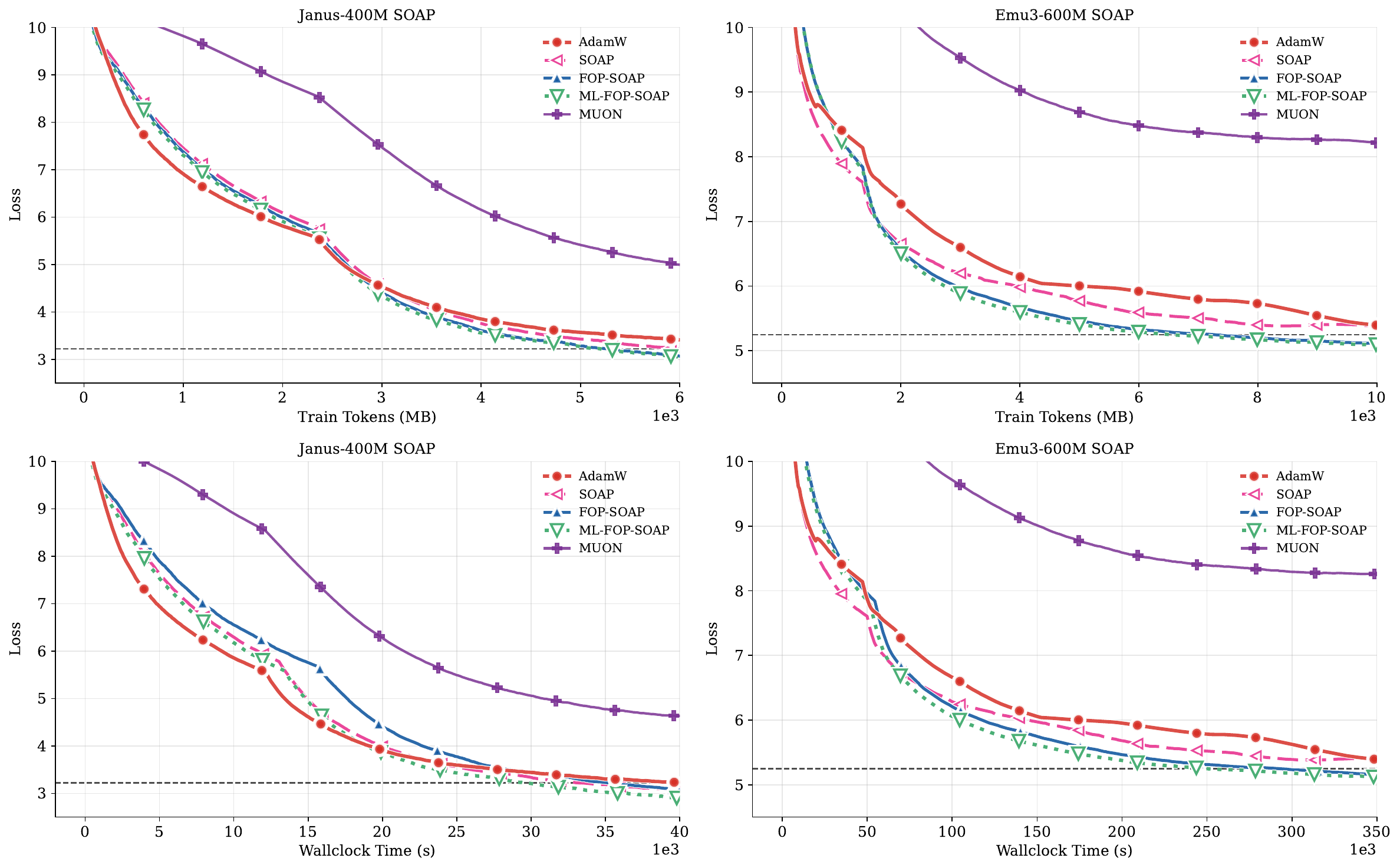}
    \caption{Training loss convergence across Janus-400M (left) and Emu3-600M (right) under a highly scaled batch size setting (Batch size = 8192). The top row tracks loss against processed Train Tokens, while the bottom row tracks it against Wallclock Time.}
    \label{fig:Janus_Emu3_LargeBatch}
\end{figure}

To explicitly verify how the proposed strategies resolve modality competition, the overall objective is decoupled into Image-to-Text (I2T) and Text-to-Image (T2I) losses in Figure~\ref{fig:Decoupled training loss curves}. In the Janus-400M architecture (left column), the global optimization trajectory is heavily dominated by the high-entropy T2I generation task, which consistently exhibits a much larger numerical magnitude than the I2T understanding task. Under this severe gradient heterogeneity, first-order methods like AdamW are overwhelmed by visual noise, resulting in sub-optimal convergence. The second-order SOAP optimizer establishes a stronger foundation by accelerating the dominant T2I task significantly faster than AdamW. However, while SOAP improves T2I generation, its corresponding I2T loss rapidly plateaus, indicating that standard second-order preconditioning still inadvertently sacrifices textual understanding to accommodate visual noise. In stark contrast, ML-FOP-SOAP successfully neutralizes this conflict; by leveraging geometry-aware variance control, it shields the fragile I2T learning process and continuously drives both losses downward simultaneously.

Furthermore, this optimization superiority is even more pronounced in the Emu3-600M architecture (right column), which operates under a pure unified next-token prediction paradigm. In this setting, the limitations of standard optimizers are drastically exposed: AdamW struggles profoundly, particularly on the I2T task, converging to a remarkably higher loss plateau than all other methods. While standard SOAP marginally improves upon AdamW, ML-FOP-SOAP achieves a strict Pareto improvement, yielding significantly deeper and faster convergence on both the I2T and T2I objectives simultaneously. These decoupled results directly substantiate the two central claims of this work: while second-order preconditioners effectively manage baseline multi-modal heterogeneity better than first-order methods, the addition of the FOP mechanism is absolutely critical for controlling variance and fully resolving modality competition across diverse architectures.

To demonstrate scalability, training dynamics are evaluated at a global batch size of 8192 (Figure~\ref{fig:Janus_Emu3_LargeBatch}). In Janus-400M and Emu3-600M, AdamW suffers from severe large-batch degradation, prematurely plateauing above baseline thresholds. While standard SOAP initially outperforms AdamW, it eventually stagnates. Only ML-FOP-SOAP maintains a continuous downward trajectory, achieving the lowest global loss. Wallclock evaluations confirm that our Multi-Level (ML) hierarchical folding strategy mitigates the computational overhead of micro-step projections, enabling robust, efficient scalability to ultra-large batches without catastrophic latency. Additional examples are provided in the supplementary materials.

\section{Conclusion}
\label{sec:conclusion}

In this work, we systematically addressed the optimization bottleneck of modality competition in natively unified multi-modal autoregressive models. We demonstrated that while the second-order SOAP optimizer establishes a robust foundation against early-stage outliers and multi-modal heterogeneity, it still falls prey to the conventional trade-off between textual understanding and visual generation. To resolve this, we introduced ML-FOP-SOAP. By flexibly applying a Fisher-Orthogonal Projection, our method actively neutralizes cross-modality variance, yielding strict Pareto improvements that drive both task objectives downward simultaneously. Furthermore, our Multi-Level hierarchical folding strategy successfully amortizes projection latency during massive gradient accumulation. This enables flawless scaling to an ultra-large batch size of 8192, breaking the late-stage convergence plateaus that plague standard optimizers. Delivering a $1.4\times$ improvement in sample efficiency and $1.5\times$ faster wall-clock convergence over AdamW, our framework provides a robust and scalable foundation for accelerating large-scale multi-modal pretraining.


%
%
\bibliographystyle{splncs04}
\bibliography{main}
%
\appendix
\clearpage
\appendix

\begin{center}
{\Large \textbf{Supplementary Material for}}\\[0.5em]
{\Large \textit{Second-Order Multi-Level Variance Correction for Modality Competition in Multimodal Models}}
\end{center}





\section{Theoretical Analysis of Fisher Information for Modality Balancing}
\label{app:theory_fisher_modality_balancing}

In this appendix, we provide a mathematical analysis of why first-order optimization can lead to \emph{modality starvation} when jointly training visual and textual modalities with heterogeneous statistical structure, and how the Fisher information matrix introduced in natural gradient descent (NGD) helps mitigate this issue through a metric transformation on the underlying Riemannian manifold.
\subsection{First-Order Dynamics of Modality Competition}
\label{app:first_order_dynamics}

Consider the parameters of a unified autoregressive multimodal model, denoted by $\theta \in \mathbb{R}^d$. In multimodal next-token prediction, the training objective is the mixed expectation of the negative log-likelihoods over image data ($I$) and text data ($T$):
\begin{equation}
\mathcal{L}(\theta)
=
\mathbb{E}_{x \sim \mathcal{D}_I}\bigl[-\log p(x \mid \theta)\bigr]
+
\mathbb{E}_{y \sim \mathcal{D}_T}\bigl[-\log p(y \mid \theta)\bigr]
\triangleq
\mathbb{E}[L_I(\theta)] + \mathbb{E}[L_T(\theta)].
\label{eq:joint_objective}
\end{equation}

In practical training, we compute the stochastic gradient using minibatches:
\begin{equation}
g = \nabla L_I(\theta) + \nabla L_T(\theta).
\label{eq:joint_gradient}
\end{equation}

Because visual tokens often carry substantially higher information density and more fine-grained local variation, whereas textual tokens lie on a more abstract semantic manifold, the corresponding empirical gradient covariance matrices are
\begin{equation}
\Sigma_I = \mathrm{Cov}(\nabla L_I(\theta)),
\qquad
\Sigma_T = \mathrm{Cov}(\nabla L_T(\theta)).
\label{eq:modality_covariances}
\end{equation}

In many settings, one empirically observes
\begin{equation}
\mathrm{Tr}(\Sigma_I) \gg \mathrm{Tr}(\Sigma_T).
\label{eq:cov_trace_imbalance}
\end{equation}

When using first-order optimizers such as AdamW, the parameter update $\Delta \theta_{\mathrm{1st}}$ is driven by exponential moving averages of the stochastic gradients. In Euclidean space, the covariance of the total stochastic gradient is
\begin{equation}
\mathrm{Cov}(g) = \Sigma_I + \Sigma_T + 2\mathrm{Cov}(\nabla L_I(\theta), \nabla L_T(\theta)).
\label{eq:joint_gradient_covariance}
\end{equation}

The cross-covariance term $2\mathrm{Cov}(\nabla L_I(\theta), \nabla L_T(\theta))$ explicitly captures the correlation and potential interference between the modalities. If $\Sigma_I$ dominates the variance scale, and the cross-covariance indicates misalignment (i.e., gradient conflict), the parameter trajectory can become disproportionately influenced by the high-variance directions induced by the visual modality. Meanwhile, the effective update signal from the text modality may be heavily suppressed or distorted by the cross-modality noise. This severe imbalance, exacerbated by standard Euclidean optimization, leads to slower progress or even stagnation for one modality, which we refer to as \emph{modality starvation}.

\subsection{KL Divergence and the Fisher Information Matrix}
\label{app:kl_fisher}

To mitigate this variance-dominated behavior under the Euclidean metric, one may instead define optimization steps on the Riemannian manifold induced by the model distribution. Let the change in model predictions induced by a parameter perturbation $\Delta \theta$ be measured by the Kullback--Leibler (KL) divergence
\begin{equation}
D_{\mathrm{KL}}(p_\theta \,\|\, p_{\theta+\Delta\theta}).
\label{eq:kl_definition_symbolic}
\end{equation}

Applying a second-order Taylor expansion of the log-likelihood around $\theta$, we obtain
\begin{equation}
\log p(z \mid \theta+\Delta\theta)
\approx
\log p(z \mid \theta)
+
\nabla \log p(z \mid \theta)^\top \Delta\theta
+
\frac{1}{2}\Delta\theta^\top \nabla^2 \log p(z \mid \theta)\Delta\theta.
\label{eq:taylor_loglikelihood}
\end{equation}

Substituting this expansion into the definition of the KL divergence yields
\begin{equation}
D_{\mathrm{KL}}(p_\theta \,\|\, p_{\theta+\Delta\theta})
=
\mathbb{E}_{p_\theta}
\left[
\log p(z \mid \theta) - \log p(z \mid \theta+\Delta\theta)
\right],
\label{eq:kl_definition}
\end{equation}
and therefore
\begin{equation}
D_{\mathrm{KL}}(p_\theta \,\|\, p_{\theta+\Delta\theta})
\approx
-\mathbb{E}_{p_\theta}\!\left[\nabla \log p(z \mid \theta)^\top \Delta\theta\right]
-\frac{1}{2}
\mathbb{E}_{p_\theta}\!\left[\Delta\theta^\top \nabla^2 \log p(z \mid \theta)\Delta\theta\right].
\label{eq:kl_taylor_expansion}
\end{equation}

Since the score function has zero expectation,
\begin{equation}
\mathbb{E}_{p_\theta}\bigl[\nabla \log p(z \mid \theta)\bigr] = 0,
\label{eq:score_zero_mean}
\end{equation}
the first term vanishes. The remaining second-order term defines the Fisher information matrix:
\begin{equation}
F
=
-\mathbb{E}_{p_\theta}\!\left[\nabla^2 \log p(z \mid \theta)\right].
\label{eq:fisher_hessian_form}
\end{equation}

Hence, the KL divergence admits the following local quadratic approximation:
\begin{equation}
D_{\mathrm{KL}}(p_\theta \,\|\, p_{\theta+\Delta\theta})
\approx
\frac{1}{2}\Delta\theta^\top F \Delta\theta
=
\frac{1}{2}\|F^{1/2}\Delta\theta\|_2^2.
\label{eq:kl_local_quadratic}
\end{equation}

This shows that, locally, the Fisher information matrix defines the intrinsic metric of the statistical manifold.

\subsection{Why Can $F^{-1}$ Mitigate Modality Competition?}
\label{app:fisher_inverse_modality}

The key observation follows from the information matrix equality (also known as Bartlett's identity). For log-likelihood objectives,
\begin{equation}
F
=
-\mathbb{E}_{p_\theta}\!\left[\nabla^2 \log p(z \mid \theta)\right]
=
\mathbb{E}_{p_\theta}\!\left[
\nabla \log p(z \mid \theta)\nabla \log p(z \mid \theta)^\top
\right].
\label{eq:bartlett_identity}
\end{equation}

Thus, the Fisher matrix simultaneously encodes both the local curvature of the loss landscape and the uncentered second moment of the score function. Under natural gradient descent, the update direction becomes
\begin{equation}
d_{\mathrm{NGD}} = F^{-1} g = F^{-1}(\nabla L_I + \nabla L_T).
\label{eq:ngd_update}
\end{equation}

To understand how this preconditioned update addresses modality imbalance, we analyze the magnitude of the update in Fisher geometry. Rather than relying on the centered covariance matrix (which strictly equals the Fisher matrix only when the expected gradient is zero at a local optimum), we examine the expected squared Fisher norm of the natural gradient directly:
\begin{equation}
\mathbb{E}\bigl[\|d_{\mathrm{NGD}}\|_F^2\bigr]
=
\mathbb{E}\bigl[d_{\mathrm{NGD}}^\top F d_{\mathrm{NGD}}\bigr]
=
\mathbb{E}\bigl[g^\top F^{-1} F F^{-1} g\bigr]
=
\mathbb{E}\bigl[g^\top F^{-1} g\bigr].
\label{eq:fisher_norm_ngd}
\end{equation}

Using the cyclic property of the trace, this can be rewritten in terms of the uncentered second moment matrix $\mathbb{E}[gg^\top]$:
\begin{equation}
\mathbb{E}\bigl[g^\top F^{-1} g\bigr]
=
\mathrm{Tr}\!\left(F^{-1}\mathbb{E}[gg^\top]\right).
\label{eq:trace_identity}
\end{equation}

In practice, the empirical Fisher information matrix is approximated precisely by this uncentered second moment of the mini-batch gradients, i.e., $F \approx \mathbb{E}[gg^\top]$. Under this standard approximation, we obtain:
\begin{equation}
\mathbb{E}\bigl[\|d_{\mathrm{NGD}}\|_F^2\bigr]
\approx
\mathrm{Tr}(F^{-1}F)
=
\mathrm{Tr}(I)
=
d,
\label{eq:fisher_norm_dimension}
\end{equation}
where $d$ is the dimension of the parameter space. 

The derivation above demonstrates a powerful property: after preconditioning by $F^{-1}$, the overall update magnitude is effectively whitened and bounded by the dimension $d$ under the intrinsic geometry of the model. Regardless of how severe the variance imbalance ($\Sigma_I \gg \Sigma_T$) or the cross-modality interference ($2\mathrm{Cov}(\nabla L_I, \nabla L_T)$) may be in the raw Euclidean gradient $g$, the natural gradient normalizes these updates. Directions associated with different modalities are no longer dominated by pure gradient scale; instead, they are structurally scaled by a metric that inherently accounts for both uncertainty and curvature. This provides a rigorous explanation for why Fisher-aware optimization fundamentally alleviates modality starvation.

\section{Formulation of Fisher-Orthogonal Projection}
\label{app:fop_derivation}

In this section, we provide the complete mathematical construction of the Fisher-Orthogonal Projection (FOP) briefly introduced in Section~\ref{sec:fop_variance_control}.

\subsection{Fisher-Orthogonal Projection (FOP)}
Given two stochastic gradients $g_1$ and $g_2$ within a mini-batch or gradient accumulation window, we define their average and difference as:
\begin{equation}
g_{\mathrm{avg}} = \frac{1}{2}(g_1 + g_2), \qquad g_{\mathrm{diff}} = g_1 - g_2
\label{eq:app_avg_diff_pair}
\end{equation}

To project $g_{\mathrm{diff}}$ onto the subspace orthogonal to $g_{\mathrm{avg}}$ under the Fisher metric, we need to compute inner products of the form $\langle u, Fv \rangle$. To avoid the intractable $\mathcal{O}(N^3)$ cost of explicitly forming $F$, we utilize the Kronecker factors maintained by the SOAP/Shampoo preconditioners to construct a forward metric proxy $\mathcal{M}(\cdot)$:
\begin{equation}
\mathcal{M}(v) \approx Fv
\label{eq:app_forward_metric_proxy}
\end{equation}

Using this efficient proxy, the Fisher-space projection coefficient is calculated as:
\begin{equation}
s_{\mathrm{proj}} = \frac{\langle g_{\mathrm{diff}}, \mathcal{M}(g_{\mathrm{avg}}) \rangle}{\langle g_{\mathrm{avg}}, \mathcal{M}(g_{\mathrm{avg}}) \rangle + \epsilon}
\label{eq:app_fisher_projection_coeff}
\end{equation}
where $\epsilon$ is a small numerical stabilizer.

We then construct the Fisher-orthogonal residual by subtracting the parallel component:
\begin{equation}
g_{\mathrm{diff}}^{\perp} = g_{\mathrm{diff}} - s_{\mathrm{proj}} g_{\mathrm{avg}}
\label{eq:app_fisher_orthogonal_residual}
\end{equation}
By construction, this residual strictly satisfies the Fisher orthogonality condition $\langle g_{\mathrm{diff}}^{\perp}, F g_{\mathrm{avg}} \rangle = 0$, ensuring that our variance correction captures pure heterogeneity without interfering with the primary descent trajectory.

Finally, the combined direction $g_{\mathrm{comb}}$ is formed and preconditioned to obtain the actual parameter update $d_{\mathrm{FOP}}$:
\begin{equation}
g_{\mathrm{comb}} = g_{\mathrm{avg}} + \beta g_{\mathrm{diff}}^{\perp}
\label{eq:app_combined_direction}
\end{equation}
\begin{equation}
d_{\mathrm{FOP}} = \mathcal{P}(g_{\mathrm{comb}}) \approx F^{-1} g_{\mathrm{comb}}
\label{eq:app_fop_update}
\end{equation}

\section{Quadratic Surrogate Analysis of Fisher-Orthogonal Projection}
\label{app:quadratic_surrogate}

In multimodal optimization, analyzing a single averaged surrogate objective often masks the underlying modality competition. To rigorously demonstrate why Fisher-Orthogonal Projection (FOP) is necessary, we model the local loss reduction using second-order Taylor approximations for the individual modalities (e.g., Vision and Text), denoted as $J_1(d)$ and $J_2(d)$.

Let the stochastic gradients of the two modalities be $g_1$ and $g_2$. We define the batch average and difference as:
\begin{equation}
g_{\mathrm{avg}} = \frac{1}{2}(g_1 + g_2), \qquad g_{\mathrm{diff}} = g_1 - g_2
\end{equation}
Consequently, individual gradients can be expressed as $g_1 = g_{\mathrm{avg}} + \frac{1}{2}g_{\mathrm{diff}}$ and $g_2 = g_{\mathrm{avg}} - \frac{1}{2}g_{\mathrm{diff}}$. The local quadratic surrogate for modality $i$ under the true Fisher information matrix $F$ is:
\begin{equation}
J_i(d) = -g_i^{\top} d + \frac{1}{2} d^{\top} F d
\label{eq:app_individual_surrogate}
\end{equation}

\subsection{Objective Value for Pure SOAP}
A standard preconditioned optimizer like SOAP targets the average gradient, applying the update $d_{\mathrm{SOAP}} = F^{-1} g_{\mathrm{avg}}$. Substituting this into the surrogate for modality 2:
\begin{align}
J_2(d_{\mathrm{SOAP}}) &= -g_2^{\top} (F^{-1} g_{\mathrm{avg}}) + \frac{1}{2} (F^{-1} g_{\mathrm{avg}})^{\top} F (F^{-1} g_{\mathrm{avg}}) \nonumber \\
&= -\left(g_{\mathrm{avg}} - \frac{1}{2} g_{\mathrm{diff}}\right)^{\top} F^{-1} g_{\mathrm{avg}} + \frac{1}{2} g_{\mathrm{avg}}^{\top} F^{-1} g_{\mathrm{avg}} \nonumber \\
&= -\frac{1}{2} \|g_{\mathrm{avg}}\|^2_{F^{-1}} + \frac{1}{2} g_{\mathrm{diff}}^{\top} F^{-1} g_{\mathrm{avg}}
\label{eq:app_j2_soap}
\end{align}
This equation perfectly illustrates \emph{modality starvation}. The term $\frac{1}{2} g_{\mathrm{diff}}^{\top} F^{-1} g_{\mathrm{avg}}$ acts as a gradient conflict penalty. If this term is positive and large, modality 2 suffers from severe optimization degradation under the shared SOAP update.

\subsection{Objective Value for FOP}
The FOP update introduces an orthogonal variance correction to break this rigid average: $d_{\mathrm{FOP}} = F^{-1}(g_{\mathrm{avg}} + \beta g_{\mathrm{diff}}^{\perp})$. Substituting this into the surrogate for the starving modality 2:
\begin{align}
J_2(d_{\mathrm{FOP}}) &= -g_2^{\top} F^{-1} (g_{\mathrm{avg}} + \beta g_{\mathrm{diff}}^{\perp}) + \frac{1}{2} (g_{\mathrm{avg}} + \beta g_{\mathrm{diff}}^{\perp})^{\top} F^{-1} (g_{\mathrm{avg}} + \beta g_{\mathrm{diff}}^{\perp}) \nonumber \\
&= -\left(g_{\mathrm{avg}} - \frac{1}{2} g_{\mathrm{diff}}\right)^{\top} F^{-1} (g_{\mathrm{avg}} + \beta g_{\mathrm{diff}}^{\perp}) \nonumber \\
& \qquad \qquad + \frac{1}{2} \left( \|g_{\mathrm{avg}}\|^2_{F^{-1}} + 2\beta g_{\mathrm{avg}}^{\top} F^{-1} g_{\mathrm{diff}}^{\perp} + \beta^2 \|g_{\mathrm{diff}}^{\perp}\|^2_{F^{-1}} \right)
\label{eq:app_j2_fop_expand}
\end{align}
Expanding the linear term yields:
\begin{equation*}
-\|g_{\mathrm{avg}}\|^2_{F^{-1}} - \beta g_{\mathrm{avg}}^{\top} F^{-1} g_{\mathrm{diff}}^{\perp} + \frac{1}{2} g_{\mathrm{diff}}^{\top} F^{-1} g_{\mathrm{avg}} + \frac{1}{2} \beta g_{\mathrm{diff}}^{\top} F^{-1} g_{\mathrm{diff}}^{\perp}
\end{equation*}
Notice the crucial algebraic cancellation of the cross-terms: the $-\beta g_{\mathrm{avg}}^{\top} F^{-1} g_{\mathrm{diff}}^{\perp}$ from the linear part perfectly cancels the $+\beta g_{\mathrm{avg}}^{\top} F^{-1} g_{\mathrm{diff}}^{\perp}$ from the quadratic part. Recombining the remaining terms gives:
\begin{align}
J_2(d_{\mathrm{FOP}}) &= \left(-\frac{1}{2} \|g_{\mathrm{avg}}\|^2_{F^{-1}} + \frac{1}{2} g_{\mathrm{diff}}^{\top} F^{-1} g_{\mathrm{avg}}\right) + \frac{1}{2} \beta g_{\mathrm{diff}}^{\top} F^{-1} g_{\mathrm{diff}}^{\perp} + \frac{1}{2} \beta^2 \|g_{\mathrm{diff}}^{\perp}\|^2_{F^{-1}} \nonumber \\
&= J_2(d_{\mathrm{SOAP}}) + \frac{1}{2} \beta g_{\mathrm{diff}}^{\top} F^{-1} g_{\mathrm{diff}}^{\perp} + \frac{1}{2} \beta^2 \|g_{\mathrm{diff}}^{\perp}\|^2_{F^{-1}}
\label{eq:app_j2_fop_final}
\end{align}

\subsection{Strict Loss Reduction for the Starving Modality}
To find the optimal adjustment that assists the starving modality, we minimize $J_2(d_{\mathrm{FOP}})$ with respect to $\beta$. Taking the derivative and setting it to zero:
\begin{equation}
\frac{\partial J_2(d_{\mathrm{FOP}})}{\partial \beta}
=
\frac{1}{2} g_{\mathrm{diff}}^{\top} F^{-1} g_{\mathrm{diff}}^{\perp}
+
\beta \|g_{\mathrm{diff}}^{\perp}\|^2_{F^{-1}}
=
0
\label{eq:app_beta_derivative}
\end{equation}
Solving for the optimal coefficient $\beta^*$:
\begin{equation}
\beta^*
=
- \frac{1}{2} \frac{g_{\mathrm{diff}}^{\top} F^{-1} g_{\mathrm{diff}}^{\perp}}{\|g_{\mathrm{diff}}^{\perp}\|^2_{F^{-1}}}
\label{eq:app_optimal_beta}
\end{equation}
Substituting $\beta^*$ back into Equation~\eqref{eq:app_j2_fop_final} yields the final loss reduction:
\begin{align}
J_2(d_{\mathrm{FOP}}(\beta^*))
&=
J_2(d_{\mathrm{SOAP}})
-
\frac{1}{8} \frac{\left(g_{\mathrm{diff}}^{\top} F^{-1} g_{\mathrm{diff}}^{\perp}\right)^2}{\|g_{\mathrm{diff}}^{\perp}\|^2_{F^{-1}}}
\label{eq:app_strict_loss_reduction}
\end{align}
Since the Fisher metric $F^{-1}$ is positive definite, the norm $\|g_{\mathrm{diff}}^{\perp}\|^2_{F^{-1}}$ is strictly positive. As long as the projected variance difference correlates with the true gradient conflict, the subtracted term is strictly positive, yielding:
\begin{equation}
J_2(d_{\mathrm{FOP}}(\beta^*)) < J_2(d_{\mathrm{SOAP}})
\end{equation}

\subsection{Physical Interpretation}
This rigorous derivation completely reframes the utility of FOP. Mathematically, a pure SOAP step based on $g_{\mathrm{avg}}$ is the strict global minimum of the \emph{average} quadratic surrogate. However, optimizing only for the average inherently punishes individual modalities with conflicting curvatures. 

Equation~\eqref{eq:app_strict_loss_reduction} algebraically proves that by intelligently injecting Fisher-orthogonal variance, FOP escapes the rigid minimum of the average surrogate to secure a strictly guaranteed loss reduction for the disadvantaged (starving) modality. FOP achieves true \emph{modality balancing} not by contradicting the Newton direction, but by dynamically correcting the modality-specific penalty induced by gradient averaging.

\section{Detailed Model Configurations and Scaling}
\label{app:model_configurations}

In Section~\ref{sec:model_and_dataset}, we introduced the scaled-down variants of the Janus and Emu3 architectures used in our experiments. Here, we provide the exact architectural hyperparameters and parameter counts to ensure full reproducibility.

\subsection{Janus Scaled Variants}
For the Janus family, we base our architectural blueprint on the official \texttt{Janus-Pro-1B} skeleton but apply structured scaling rules to the number of transformer layers, hidden dimensions, and attention heads. The multimodal projection layers scale proportionally with the language model's hidden size.

\begin{itemize}
    \item \textbf{Janus-400M :} Configured with 14 hidden layers, a hidden dimension of 1024, an intermediate FFN size of 2816, and 8 attention heads. This configuration yields approximately 408 million trainable parameters (with a total of 799 million parameters when including frozen vision embeddings).
    \item \textbf{Janus-1B:} Configured with 20 hidden layers, a hidden dimension of 1536, an intermediate FFN size of 4096, and 12 attention heads. This configuration yields approximately 911 million trainable parameters (with a total of 1.30 billion parameters).
\end{itemize}

\subsection{Emu3 Scaled Variants}
For the Emu3 family, we utilize explicit configuration presets to shrink the model scale while strictly maintaining the core Emu3 architectural properties and vocabulary size (184,622).

\begin{itemize}
    \item \textbf{Emu3-600M:} Configured with 32 hidden layers, a hidden dimension of 1024, an intermediate FFN size of 2816, and 16 attention heads (with 16 KV heads). This configuration yields an active parameter count of approximately 600 million.
\end{itemize}

By utilizing these precise configurations, we maintain the fundamental multimodal optimization challenges of the original architectures while making large-batch gradient accumulation studies computationally feasible.

\subsection{Hyperparameter Search and Reproducibility}
\label{app:hyperparameter_search}

To guarantee the exact reproducibility of our optimization trajectories, all experiments—including baselines and our proposed methods—are conducted using a fixed random seed of $0$. 

Given the sensitivity of second-order and preconditioned optimizers to the learning rate, we refrain from using heuristic defaults. Instead, we perform a rigorous half-decade logarithmic grid search to determine the optimal peak learning rate for each configuration. The search space is defined by the set:
\begin{equation}
\eta \in \{0.1, 0.0316, 0.01, 0.00316, 0.001, 0.000316\}
\label{eq:lr_grid}
\end{equation}
which perfectly corresponds to steps of $10^{-0.5}$ in the logarithmic scale. 

For the tensor preconditioners (Shampoo and SOAP) imported from the \texttt{distributed\_shampoo} repository, we retain their official default structural hyperparameters (e.g., block sizes and preconditioner update frequencies) to ensure that the baselines operate at their intended optimal algorithmic capacity.

\end{document}